\documentclass[conference]{IEEEtran}
\IEEEoverridecommandlockouts
\usepackage{amsmath,amsfonts}
\usepackage{algorithmic}
\usepackage{algorithm}
\usepackage{array}
\usepackage[caption=false,font=normalsize,labelfont=sf,textfont=sf]{subfig}
\usepackage{textcomp}
\usepackage{stfloats}
\usepackage{url}
\usepackage{verbatim}
\usepackage{graphicx}
\usepackage{booktabs}
\usepackage{multirow}
\usepackage{tabularx}
\usepackage{tabulary,xcolor}
\usepackage{cite}
\usepackage{soul}
\usepackage{xspace}
\usepackage{enumitem}

\def\BibTeX{{\rm B\kern-.05em{\sc i\kern-.025em b}\kern-.08em
    T\kern-.1667em\lower.7ex\hbox{E}\kern-.125emX}}
\begin{document}
\newcommand{\model}{SolarFormer\xspace}

\title{\huge{SolarFormer: Multi-scale Transformer for Solar PV Profiling}\\
\thanks{This material is based upon work supported by the National Science Foundation (NSF) under Award OIA-1946391 RII Track-1, NSF 2119691 AI SUSTEIN, NSF 2236302}}

\author{\IEEEauthorblockN{1\textsuperscript{st} Adrian De Luis$^*$, 2\textsuperscript{nd} Minh Tran$^*$, 3\textsuperscript{nd} Taisei Hanyu, 4\textsuperscript{nd} Anh Tran}
\IEEEauthorblockA{\textit{AICV Lab, Dept.  EECS} \\
\textit{Fayetteville, AR 72701 USA}\\
\{ad084, minht, thanyu, anhtran\}@uark.edu\\
}
\and
\IEEEauthorblockN{5\textsuperscript{rd} Liao Haitao}
\IEEEauthorblockA{\textit{Dept.  Industrial Eng.} \\
\textit{Fayetteville, AR 72701 USA}\\
liao@uark.edu\\
}
\and
\IEEEauthorblockN{6\textsuperscript{rd} Roy McCann, 7\textsuperscript{rd} Alan Mantooth}
\IEEEauthorblockA{\textit{Dept.  EECS} \\
\textit{Fayetteville, AR 72701 USA}\\
\{rmccann, mantooth\}@uark.edu\\
}
\and
\IEEEauthorblockN{8\textsuperscript{th} Ying Huang}
\IEEEauthorblockA{\textit{Civil, Cons. and Env. Eng.} \\
\textit{Fargo, ND 58102 USA}\\
ying.huang@ndsu.edu}
\and
\IEEEauthorblockN{9\textsuperscript{th} Ngan Le}
\IEEEauthorblockA{\textit{AICV Lab, Dept.  EECS} \\
\textit{Fayetteville, AR 72701 USA}\\
thile@uark.edu\\
}}
\maketitle

\begin{abstract}

As climate change intensifies, the global imperative to shift towards sustainable energy sources becomes more pronounced. Photovoltaic (PV) energy is a favored choice due to its reliability and ease of installation. Accurate mapping of PV installations is crucial for understanding their adoption and informing energy policy. To meet this need, we introduce the \model, designed to segment solar panels from aerial imagery, offering insights into their location and size. However, solar panel identification in Computer Vision is intricate due to various factors like weather conditions, roof conditions, and Ground Sampling Distance (GSD) variations. To tackle these complexities, we present the \model, featuring a multi-scale Transformer encoder and a masked-attention Transformer decoder. Our model leverages low-level features and incorporates an instance query mechanism to enhance the localization of solar PV installations. We rigorously evaluated our \model using diverse datasets, including GGE (France), IGN (France), and USGS (California, USA), across different GSDs. Our extensive experiments consistently demonstrate that our model either matches or surpasses state-of-the-art models, promising enhanced solar panel segmentation for global sustainable energy initiatives. Source code is available upon acceptance.

\end{abstract}

\begin{IEEEkeywords}
Aerial Images, Solar PVs, Profiling, Segmentation, multi-scale, Transformers.
\end{IEEEkeywords}

\section{Introduction}
The growth of photovoltaic (PV) energy production has been impressive, driven by improved cell manufacturing \cite{PETERS20192732}, a competitive market, and increasing energy costs \cite{LEMAY2023113571}. In 2022, the U.S. Energy Information Administration reported a 440\% increase in small-scale PV installation capacity, reaching 39.5 GW from 7.3 GW in 2014 \cite{EIA2022}.

While solar panels offer sustainability, they pose challenges for energy forecasting \cite{PIERRO2022983}. PV energy relies on weather conditions, operates in decentralized systems, and often remains unseen by Transmission System Operators \cite{7347457}. Accurate forecasting is crucial, especially in high PV adoption areas, to prevent grid congestion and imbalances \cite{PIERRO2022983}.

To gain real-time insights into PV distribution and energy production, recent research has focused on profiling PV installations using remote sensing imagery from unmanned vehicles \cite{YU20182605} and satellites, coupled with Machine Learning and Computer Vision. Remote Sensing (RS) images provide an overhead view of the Earth's surface and define the level of detail as Ground Sample Distance (GSD). RS imagery serves multiple practical purposes, including climate change analysis \cite{ONEILL2013413}, natural disaster monitoring \cite{DBLP:journals/corr/abs-1901-04277}, urban planning \cite{9324098}\cite{Effat2013AMA}, and land cover analysis \cite{NEURIPS_DATASETS_AND_BENCHMARKS2021_4e732ced}.

\begin{figure*}[!t]
\centering
\includegraphics[width=0.9\linewidth]{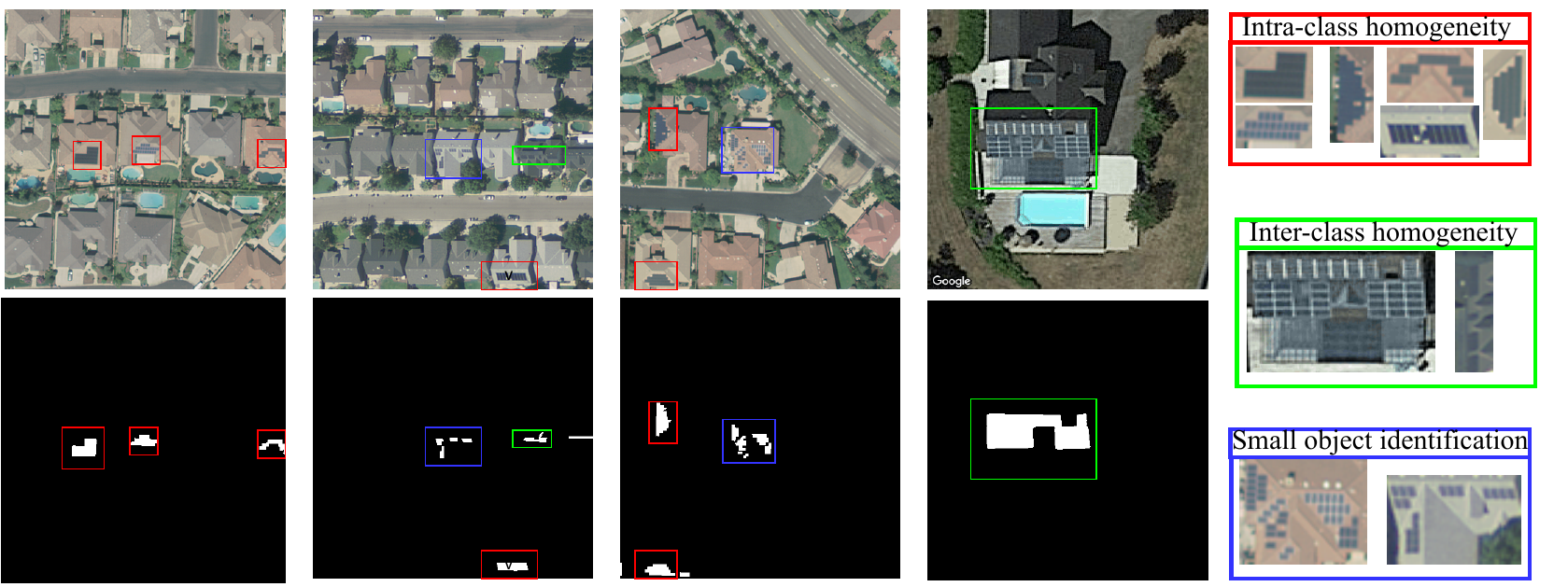}
\caption{Examples of challenging characteristics of solar PV segmentation. Within a class, there is a large diversity in appearance: intra-class heterogeneity (red); some different classes share the similar appearance: inter-class homogeneity (green), solar PV are dense and small such that they are hardly identifiable (blue)).}
\label{fig:solar_pv_challenges}
\end{figure*}

Deep Learning (DL) has achieved remarkable image understanding capabilities, even surpassing human performance. DL has been applied to extract solar PV distribution through a two-step process: training a classifier to identify images with solar panels and passing them to the segmentation network such as Kasmi, et al., \cite{kasmi2023unsupervised}, HyperionSolarNet\cite{parhar2022hyperionsolarnet}. However, existing DL-based methods have two major limitations: (i) They employ a two-stage framework, consisting of separate classifier and segmentor networks. This approach heavily relies on the classifier network, leading to suboptimal learning in the segmentation network. Essentially, they adapt existing DL frameworks designed for natural images to solar PV imagery, neglecting the unique challenges posed by the latter. (ii) These existing DL frameworks are originally designed for natural images and do not adequately address the specific challenges encountered in solar PV imagery. As depicted in Fig.\ref{fig:solar_pv_challenges}, these challenges include intra-class heterogeneity, inter-class homogeneity, and the identification of small objects.

To overcome these limitations, we have harnessed recent advances in Deep Learning, particularly the Transformer architecture \cite{vaswani2017attention}, to introduce a novel model called \model. Our \model incorporates a multi-scale Transformer encoder and a masked-attention Transformer decoder to address these challenges effectively. Our contributions can be summarized as follows:

\begin{itemize}
    \item Implementation of a high-resolution Transformer to enhance solar PV understanding, particularly in cases with low resolution and GSD variations.
    \item Deployment of a multi-scale Transformer encoder with a masked-attention Transformer decoder to extract rich visual representations, effectively addressing inter-class and intra-class homogeneity challenges.
    \item An extensive comparative evaluation between our \model and existing DL models using three public datasets, demonstrating the effectiveness of our proposed approach.
\end{itemize}

\section{Related Works}

\subsection{Deep Learning-based Solar PV Analysis}

  
In recent times, the mapping of high-fidelity solar installations has become increasingly important. This uptick in interest is a direct result of the rapid adoption of photovoltaic (PV) energy, enhanced resolution of aerial imagery, and the advancements in DL techniques. The preliminary efforts in this field focused on binary classification of images, determining the presence or absence of PV installations \cite{MALOF2016229} \cite{8127092}. 

DeepSolar \cite{YU20182605} utilizes a Convolutional Neural Network (CNN) as a dual-stage network which includes classification and segmentation strategy, marked a significant milestone in solar PV profiling. The success of this endeavor paved the way for refining the characterization network. This was achieved by integrating state-of-the-art (SOTA) CNN methods into classifier and segmentation architectures. Meanwhile, certain researchers opted to concentrate solely on segmentation such as Zhuang, et al., \cite{ZHUANG2020106283}, Zech, et al., \cite{9300636}. Furthermore, these methodologies are adaptable for generating PV capacity estimates 3D-PV-Locator\cite{MAYER2022118469}, Kasmi, et al., \cite{kasmi2023unsupervised} and conducting socioeconomic analyses such as DeepSolar \cite{YU20182605}. Such assessments leverage the dimensions and geographical data of the segmented installations 

A prominent challenge is the creation of suitable datasets. The complications arise from (i) the extensive need for annotated images, a process that's both time-consuming and meticulous, and (ii) the potential sensitivity of the data, making it unsuitable for public distribution. Fortunately, organizations like IGN and USGS have been proactive, offering freely accessible aerial images. Despite this support, researchers often find themselves in a bind — they either annotate a limited selection of aerial imagery such as Hyperion-Solar-Net\cite{parhar2022hyperionsolarnet}, keep the collected data unreleased or organize crowd sourced annotating campaigns such as IGN-France \cite{Kasmi_2023}, USGS, California  \cite{California2016}.

\subsection{Transformer}
Transformer \cite{vaswani2017attention} and Vision Transformer (ViT) \cite{dosovitskiy2020image} have recently attracted significant interest in the research community. Initially implemented for Natural Language Processing (NLP) tasks, the Transformer \cite{vaswani2017attention} incorporates a self-attention mechanism to enhance learning on long-range relationships inside the data. It also can be adapted for parallelization, allowing for faster training on multiple nodes for large datasets. This capabilities have proven very useful for multiple NLP tasks \cite{9222960}. 
The ViT divides a 2D image into patches and then feeds the 1D sequence of linear embedding into the Transformer in a similar manner as how words would be treated for a NLP Transformer. This scheme proved specially useful when dealing with large datasets, where the ViT was able to attend to the information from the lowest levels and produce better global context than CNNs. This novel approach has already been implemented for a variety of tasks such as image recognition \cite{touvron2021training, joo2023clip}, video captioning \cite{yamazaki2022vlcap} \cite{yamazaki2023vltint}, action localization \cite{vo2022contextual, vo2023aoe}, aerial imaging \cite{kasmi2023unsupervised} object detection \cite{sun2021sparse, tran2022aisformer} and image segmentation \cite{tran2022aisformer}, medical imaging \cite{nguyen2023embryosformer} proving their capacity to synthesise global information.

Presently, most DL-based methods for solar PV analysis are merely adaptations of pre-existing DL techniques tailored for solar PV data. Consequently, they frequently fail to tackle unique challenges inherent to solar PV datasets, such as low resolution, intra-class variability, and inter-class overlap. In this paper, we introduce a multi-scale high-resolution Transformer network specifically designed to extract solar PV data while effectively addressing the aforementioned challenges. Inspired by the recent advances in DL and CV with the ViT, we incorporate the use of the Transformer for the task of semantic segmentation and propose a new model \model.

\begin{figure*}[!t]
\centering
\includegraphics[width=\linewidth]{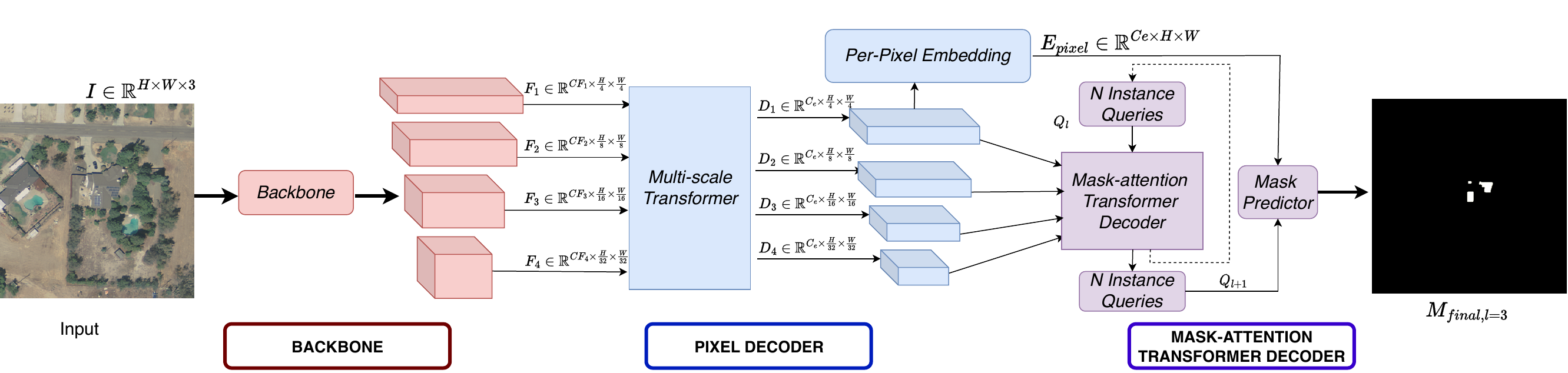}
\caption{Overall network architecture of our proposed \model which consists of three components i.e., Backbone, Pixel Decoder, and Mask-attention Transformer Decoder.}
\label{network_arch}
\end{figure*}

\section{Methods}



This section provides a detailed description of our transformer-based framework \model. This framework is specifically tailored for the segmentation of PV installations. As illustrated in Fig.\ref{network_arch}, our network comprises three distinct components: Network Backbone, Pixel Decoder, Mask-attention Transformer Decoder. Given an RGB image with dimensions $H \times W \times 3$, the network produces a corresponding segmentation mask of dimensions $H \times W$.


\subsection{Network Backbone}

The first block of our architecture is represented by the Backbone networks, which extract features from the input image of dimensions $H \times W \times 3$ to be further processed by the downstream architecture. Backbone networks are crucial in network design, and various implementations can be found in the literature. VGG \cite{simonyan2015deep}, for instance, was developed as a deep CNN suitable for image classification, boasting a depth of up to 19 layers. However, its straightforward design, primarily consisting of pooling and fully connected layers, paved the way for other CNN backbones to rise to prominence.
CNN backbones include ShuffleNet \cite{DBLP:journals/corr/ZhangZLS17}, Inception \cite{szegedy2014going}, DenseNet \cite{DBLP:journals/corr/HuangLW16a}, EfficientNet \cite{DBLP:journals/corr/abs-1905-11946}, etc. One of the most popular backbone networks are Residual Networks (ResNets) \cite{DBLP:journals/corr/HeZRS15}. This family of networks is based on blocks of convolution and pooling layers with skipped connections and recurrent units between them followed by batch normalization's. The ResNet family has a variety of implementations: ResNet-50, ResNet-34, ResNet-101, etc. and has been widely used for image classification and semantic segmentation tasks. For our model, we will be using ResNet pretrained on ImageNet as the backbone for our implementation and testing the performance of two of its most popular varieties: ResNet-50 and ResNet-101.


By employing ResNet as our proposed model's backbone, an input image $I \in \mathbb{R}^{H\times W \times 3}$ is transformed into several multi-scale feature maps $F$. In our model, we produce feature maps at multiple resolutions. Specifically, we generate four such feature maps denoted as $F_1 \in \mathbb{R}^{C_{F_1}\times \frac{H}{4} \times \frac{W}{4}}$, $F_2 \in \mathbb{R}^{C_{F_2}\times \frac{H}{8} \times \frac{W}{8}}$, $F_3 \in \mathbb{R}^{C_{F_3}\times \frac{H}{16} \times \frac{W}{16}}$ and $F_4 \in \mathbb{R}^{C_{F_4}\times \frac{H}{32} \times \frac{W}{32}}$, where $C_{F_i}$ represents the number of channels, and $H \times W$ is the size of the input image.

\subsection{Pixel Decoder}

The Pixel Decoder module serves as an intermediary between the Backbone Network and the Mask-Attention Transformer Decoder. It consists of two main components: the Multi-Scale Transformer Encoder and the Per-Pixel Embedding Module. The first component produces enhanced embedding features $D_1, D_2, D_3, D_4$ whereas the later component generates per-pixel embeddings for the image, denoted as $E_{pixel}$. 

\subsubsection{Multi-Scale Transformer Encoder}
\label{multiscale_tf_encoder}

This module takes the last four feature maps $F_1, F_2, F_3, F_4$ generated by the backbone network and processes them hierarchically. These feature maps are ordered from low to high resolution and flattened using an embedding projection to achieve a consistent channel size $C_e$, obtaining $S_i \in \mathbb{R}^{H_i \times W_i \times C_e}$ where $i={1,2,3,4}$ and $C_e$ is equal to the size of $C_{F_1}$. The flattened embeddings are then merged into $S \in \mathbb{R}^{K\times C_e}$, where $K = \sum_{i\in\{4,3,2,1\}} H_i \cdot W_i $. Since the embedding features $S$ are flattened out of their original spatial shapes and scale level, they are supplemented with a learnable encodings $L$ and $P$ of shape $S, P, L\in \mathbb{R}^{K\times C_e}$. The three encodings are then passed through a Multi-Scale Transformer Encoder \cite{dosovitskiy2021an} to produce learned features from the input sequence, taking the embedded features and generating encoded features that capture the relationships between the elements. The correlated feature embeddings $E_i \in \mathbb{R}^{H_i\cdot W_i\times C_e}$ are then divided into groups based on the multi-scale level and restored to their original spatial shape with a fixed channel size $C_e$ as $D_1 \in \mathbb{R}^{C_e\times \frac{H}{4} \times \frac{W}{4}}$, $D_2 \in \mathbb{R}^{C_e\times \frac{H}{8} \times \frac{W}{8}}$, $D_3 \in \mathbb{R}^{C_e\times \frac{H}{16} \times \frac{W}{16}}$ and $D_4 \in \mathbb{R}^{C_e\times \frac{H}{32} \times \frac{W}{32}}$.

\subsubsection{Per-Pixel Embedding Module}
\label{perpixel_emb_mod}

The second stage of the Pixel Decoder computes the pixel embeddings $E_{pixel}$ using the first feature layer $D_1$ outputted from the  Multi-Scale Encoder. The goal of this process is to scale the $D_1 \in \mathbb{R}^{C_e\times \frac{H}{4} \times \frac{W}{4}}$ to the original spatial shape $H \times W$ of the image and creating $E_{pixel} \in \mathbb{R}^{Ce \times H \times W}$. Each pixel in the $E_{pixel}$ represents both the semantic and mask classification of the corresponding pixel on the original image.

\subsection{Mask-attention Transformer Decoder}

\subsubsection{Mask Predictor}

To predict the segmented masks of possible instances in every image, per-pixel embeddings $E_{pixel}$ are utilised. The prediction process involves learning $N$ per-segment query embeddings $Q \in \mathbb{R}^{N \times C_e}$, where $N$ represent the features of the maximun amount of possible instances in the image. Each $Q$ then correlates with every pixel feature in $E_{pixel}$ and determines if the pixel belongs to the corresponding instance. Therefore, the predicted instance segmentation becomes $M \in \mathbb{R}^{N \times  H \times W}$.


\subsubsection{Mask-attention Transformer Decoder}

This module decodes $N$ per-segment query embeddings $Q \in \mathbb{R}^{N \times C_q}$ from the encoded feature maps $D_1, D_2, D_3, D_4$. The decoding procedure is done recurrently on different steps, where each step is associated to a layer from the encoded feature maps. At the first step $l=0$ we are processing the encoded feature layer with the lowest resolution $D_4$, and we recurrently process each layer all the way to the highest resolution $D_1$. At each of these layers, the query $Q_{l+1}$ is decoded from the previous layer's $Q_l$ and its corresponding encoded feature maps.

A predicted mask $M_l$ is subsequently computed using the current $Q_l$ query and the per-pixel embeddings $E_{pixel}$ and interpolated to the same size as $D_{4-l}$. $M$ is then used as an attention mechanism that focuses on the salient parts of the features maps and allows for the query embeddings to attend to the features that are most important to the instances being decoded. Specifically, the attention mask is applied to $D_{4-l}$ during the decoding process to selectively attend to certain areas of the feature maps most relevant to the instance decoded.

\subsubsection{Segmentation Mask Prediction}

The last step of the model generates the final segmentation masks $M_{final} \in \mathbb{R}^{N \times  H \times W}$. To do so, the query embeddings from the decoder at the last step $l=3$, $Q_{l=3}$ and the per-pixel embeddings $E_{pixel}$ are utilized to compute $M_{final}$.


\section{Experiments}

\subsection{Datasets}


Our \model is benchmarked using three distinct RGB aerial imagery datasets. Each dataset is partitioned using a 60/20/20 split for training, testing, and validation, respectively. To maintain uniformity, the total number of images featuring PV installations is equally distributed across these divisions. The annotated data primarily consists of two classes: PV installation and background. The datasets sourced were from three distinct regions:


\noindent
\textbf{USGS, California}: This aerial imagery, sourced from the United States Geological Survey (USGS), is part of their continued initiative to map various U.S. regions using detailed imagery, focusing specifically on California. Annotations are provided by  \cite{California2016}. This dataset boasts 601 TIF images, each with dimensions of 5000x5000 pixels and a resolution finer than 30 cm/pixel. Each image represents a 2.25 square kilometer area. For consistency across datasets, we extracted 400x400 patches from these images, resulting in a total of 37,660 images, of which 50.87\% feature PV installations.

\noindent 
\textbf{IGN, France}: This large high-resolution aerial imagery dataset was gathered from the publicly available Institut géographique national (IGN) website in France. Similarly to the process followed with the California dataset, these images were annotated through a crowd-sourcing campaign \cite{Kasmi_2023}. The dataset contains 17,325 thumbnails sized 400-by-400 at a resolution of 20 cm/pixel with a share of 44.34 \% containing solar installations.

\noindent \textbf{GGE, France}: While this dataset's aerial imagery overlaps with the regions covered in the IGN, France dataset, the images were specifically derived from Google Earth Engine. Boasting a GSD of 10 cm/pixel, this resolution is the most refined of all datasets used in our experiments. The same crowd-sourcing initiative as mentioned in \cite{Kasmi_2023} was employed for annotations. This dataset includes 28,807 thumbnail images, each measuring 400x400 pixels, with a ratio of 46.11\% depicting solar installations versus those that don't.

\begin{table*}[!t]
\centering
\caption{Performance comparison between our \model with existing \textbf{DL-based Solar PV profiling methods}. The \textbf{bold} and \underline{\textit{ITALIC UNDERLINE}} represent the best and second best performances.}
\setlength{\tabcolsep}{6pt}
\begin{tabular}{l|ccc|ccc|ccc}
\toprule
 \multirow{3}{*}{\textbf{ Methods }} & \multicolumn{9}{c}{\textbf{Solar PV Segmentation Performance} on each dataset} \\ \cline{2-10}
  &  \multicolumn{3}{c|}{GGE, France} & \multicolumn{3}{c|}{USGS, California} & \multicolumn{3}{c}{IGN, France}  \\ \cline{2-10}
  &  \textbf{IoU} & \textbf{F1-score} & \textbf{Accuracy} & \textbf{IoU} & \textbf{F1-score} & \textbf{Accuracy} & \textbf{IoU} & \textbf{F1-score} & \textbf{Accuracy}\\ \hline
Zech et al.,~\cite{9300636} & $68.59$ & $81.40$ & $77.79$ & $69.80$ & $73.29$  & $82.20$ & $38.60$ & $55.69$  & $45.19$ \\
3D-PV-Locator~\cite{MAYER2022118469} & \textbf{80.70} & \textbf{89.30} & \underline{\textit{90.70}} & $80.60$ & $89.30$ & $87.40$ & \underline{\textit{53.10}} & \underline{\textit{69.40}}  & \underline{\textit{66.40}}\\
Zhuang, et al.,~\cite{ZHUANG2020106283}& $76.60$ & $86.69$  & $84.20$ & \underline{\textit{84.39}} & \underline{\textit{91.60}}  & \underline{\textit{90.89}} & $48.50$ & $65.30$  & $59.60$\\
\hline
\textbf{Our \model}  & \underline{\textit{79.42}} & \underline{\textit{88.53}} &  \textbf{94.18} & \textbf{88.78}  & \textbf{94.05} & \textbf{94.39} & \textbf{56.41} & \textbf{71.13}  & \textbf{79.66} \\
 \bottomrule
\end{tabular}
\label{tab:comparison}
\end{table*}

\begin{table*}[!t]
\centering
\caption{Performance comparison between our \model with existing \textbf{DL-segmentation Networks} on various \textbf{backbones}. The \textbf{bold} and \underline{\textit{ITALIC UNDERLINE}} represent the best and second best performances for each backbone network}
\setlength{\tabcolsep}{6pt}
\begin{tabular}{l|l|ccc|ccc|ccc}
\toprule
\multirow{3}{*}{\textbf{Backbones }} & \multirow{3}{*}{\textbf{ Networks }} & \multicolumn{9}{c}{\textbf{Solar PV Segmentation Performance} on each dataset} \\ \cline{3-11}
 &  &  \multicolumn{3}{c|}{GGE, France} & \multicolumn{3}{c|}{USGS, California} & \multicolumn{3}{c}{IGN, France}  \\ \cline{3-11}
 &  &  \textbf{IoU} & \textbf{F1-score} & \textbf{Accuracy} & \textbf{IoU} & \textbf{F1-score} & \textbf{Accuracy} & \textbf{IoU} & \textbf{F1-score} & \textbf{Accuracy}\\ 
\toprule
\multirow{7}{*}{ResNet-50}  & FCN  \cite{long2015fully} & $74.1$ & $85.12$ & $86.59$ & $63.02$ & $77.32$ & $75.55$ & $45.23$ & $62.29$  & $59.97$\\
  & UNet \cite{DBLP:journals/corr/RonnebergerFB15} & $76.60$ & $86.69$  & $84.20$ & $84.39$ & $91.60$  & $90.89$ & $48.50$ & $65.30$  & $59.60$\\
  & PSPNet \cite{zhao2017pyramid} & \underline{\textit{77.79}} & \underline{\textit{87.50}} & $85.50$ & $77.30$ & $87.19$ & $86.10$ & $50.90$ & $67.50$  & $62.90$ \\
  & UperNet \cite{xiao2018unified} & \textbf{79.40} & \textbf{88.49} & $89.80$ & $84.50$ & $91.60$ & $90.49$ & $52.89$ & $69.19$  & $65.49$ \\
  & Mask2Former \cite{Cheng_2022_CVPR} & $74.14$ & $85.15$ & \underline{\textit{90.73}} & \underline{\textit{85.33}} &  \underline{\textit{92.08}}  & \textbf{92.80} & $54.09$ & $70.21$ &  \textbf{87.63}\\ \cline{2-11}
 & \textbf{Our \model}  & $77.65$ & $87.42$ &  \textbf{93.08} & \textbf{87.39} & \textbf{93.27} & \underline{\textit{92.76}} & \textbf{56.06} & \textbf{71.84}  & \underline{\textit{82.25}} \\
\midrule
\multirow{7}{*}{ResNet-101}  & FCN  \cite{long2015fully} & $73.20$ & $84.53$ & $87.14$ & $61.83$ & $76.41$ & $73.55$ & $45.52$ & $62.55$ & $62.43$ \\
 & UNet \cite{DBLP:journals/corr/RonnebergerFB15} & $68.59$ & $81.40$ & $77.79$ & $69.80$ & $73.29$  & $82.20$ & $38.60$ & $55.69$  & $45.19$\\
 & PSPNet \cite{zhao2017pyramid} & $78.29$ & $86.29$ & $87.80$ & $76.70$ & $86.79$ & $85.50$ & $48.80$ & $65.60$ & $59.10$\\
 & UperNet \cite{xiao2018unified} & \underline{\textit{79.19}} & \underline{\textit{88.40}} & $90.10$ & $83.79$ & $89.30$ & $91.20$ & \underline{\textit{52.20}} & \underline{\textit{68.59}}  & $65.10$\\

& Mask2Former \cite{Cheng_2022_CVPR} & $77.03$ & $87.02$ & \underline{\textit{92.39}} & \underline{\textit{86.98}} & \underline{\textit{93.03}} & \underline{\textit{94.10}} & $49.34$ & $66.08$  & \underline{\textit{66.52}}\\
\cline{2-11}
 & \textbf{Our \model}   & \textbf{79.42} & \textbf{88.53} &  \textbf{94.18} & \textbf{88.78}  & \textbf{94.05} & \textbf{94.39} & \textbf{56.41} & \textbf{71.13}  & \textbf{79.66} \\

\bottomrule
\end{tabular}
\label{tab:solar_PV_results}
\end{table*}

\begin{figure*}[!t]
\centering
\includegraphics[width=\linewidth]{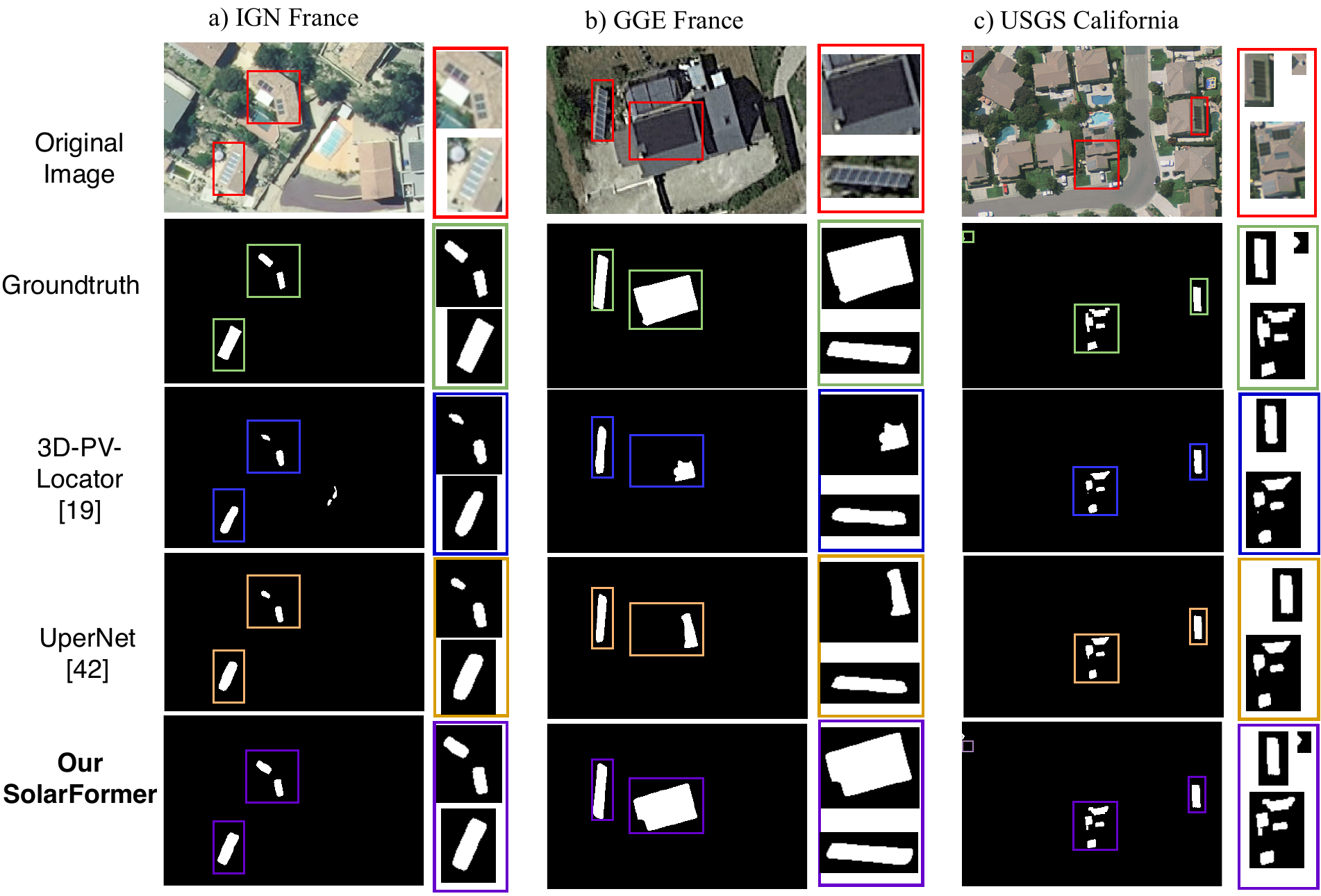}
\caption{Qualitative comparison on (a) IGN France, (b) GGE France, (c) USGS California. From top to bottom: Original RGB Image, Groundtruth, Upernet \cite{xiao2018unified} and DeepLabv3+ \cite{chen2017rethinking} and our \model.}
\label{fig:solar_pv_comparison}
\end{figure*}

\subsection{Evaluation Metrics}

To assess the efficacy of our \model and compare it with existing methods, we utilize three standard metrics often employed in semantic segmentation evaluations: Intersection over Union (IoU), F1 score, and Accuracy. These metrics hinge on the comprehension and calculation of four key values: true positive (TP), true negative (TN), false positive (FP), and false negative (FN) as follows:
\begin{equation}
    \text{IoU} = \frac{TP}{TP + FN + FP}
\end{equation}
\begin{equation}
    \text{Accuracy} = \frac{TP + TN}{TP + TN + FN + FP}
\end{equation}
\begin{equation}
    \text{F1-score} =\frac{2TP}{2TP + FN + FP}
\end{equation}








\subsection{Implementation Details}


Our implementation is based on Mask2former \cite{Cheng_2022_CVPR}. We trained \model using two RTX 8000 GPUs. We used batches of 16 images per iteration and totalled 40 epochs per dataset. We employed an Adam \cite{kingma2014adam} optimizer with a learning rate of $1\times10^{-4}$ and a weight decay of 0.05.

\subsection{Quantitative Results and Analysis}

We evaluate our \model in terms of both network performance and network design efficiency.
\textbf{i. Network Performance}: Our \model is compared with current state-of-the-art (SOTA) DL-based Solar PV profiling methods, as detailed in Table \ref{tab:comparison}.
\textbf{ii. Network Design Efficiency}: We compare our \model against other prevalent DL-based segmentation models employing various backbone networks (i.e. ResnNet-50 and ResNet-101). This comparison is presented in Table \ref{tab:solar_PV_results}. These quantitative assessments were conducted across all three datasets.

\noindent
\textbf{i. Network Performance}

Table \ref{tab:comparison} summarizes the performance of our \model against current SOTA methods for solar profiling. Our segmentation model outperforms Zech et al.,~\cite{9300636}, 3D-PV-Locator~\cite{MAYER2022118469} and Zhuang, et al.,~\cite{ZHUANG2020106283} on all three datasets. For instance, when analyzing the USGS California dataset, our model exhibits substantial improvements, including an impressive 4.39\% increase in IoU, a notable 2.45\% enhancement in F1-score, and a substantial boost of 3.5\% in Accuracy compared to the second-best method Zhuang, et al.,~\cite{ZHUANG2020106283}. Similarly, on IGN France, our model showcases remarkable advancements, with gains of 3.31\%, 1.73\%, and 13.26\% in IoU, F1-score, and Accuracy, respectively, when compared to the runner-up method, 3D-PV-Locator~\cite{MAYER2022118469}.

\noindent
\textbf{ii. Network Design Efficiency}
Table \ref{tab:solar_PV_results} presents performance comparisons between our \model and the SOTA methods. Overall, our \model achieves its highest performance using ResNet-101, while the second-best performance is delivered by our \model implemented with ResNet-50.



\noindent
\underline{\textit{GGE, France.}} Our \model demonstrates significant improvements over CNN-based models. Specifically, when benchmarked on the ResNet-101 backbone, it gains 0.23\% IoU and 0.13\% F1-score on UperNet~\cite{xiao2018unified} and improves the Accuracy result of Mask2former~\cite{Cheng_2022_CVPR} by 1.79\%. On the ResNet-50 backbone, \model outperforms the second best method, Mask2Former~\cite{Cheng_2022_CVPR}, by 2.35\% on Accuracy. Although the high precision of the 10 cm/pixel GSD imagery likely aids \model in discerning finer image details, the granularity of the images prevents greater performance gains as shown on the other datasets.


\noindent
\underline{\textit{USGS, California.}} Again, \model displays notable gains over both CNN-based models and the Mask2Former~\cite{Cheng_2022_CVPR} on both ResNet backbones. For example on the ResNet-50 backbone, when compared to the top performing CNN-based model UperNet~\cite{xiao2018unified}, \model gains 4.99\% on IoU, 4.75\% on F1-score and 3.19\% on Accuracy. The difference with Mask2former~\cite{Cheng_2022_CVPR} is still remarkable at 1.8\%, 1.02\% and 0.29\%. Despite a high GSD, California's vast aerial imagery and reduced granularity seem to enhance the model's feature recognition. 



\noindent
\underline{\textit{IGN, France.}} Despite its lower resolution, \model exhibits robust scores against other models on the ResNet backbones. The dataset's underwhelming performance is attributed to its lower resolution and limited imagery. Similar to GGE, France dataset, the granularity in the imagery might impact the model's efficiency in detecting installations. For example, when benchmarked on the ResNet-101 backbone, it achieves gains of 4.21\% on Iou and 2.54\% on the UperNet~\cite{xiao2018unified} and 13.14\% on Accuracy on the Mask2Former~\cite{Cheng_2022_CVPR}.

\subsection{Qualitative Results and Analysis}


In this section, we provide a qualitative comparison of \model against widely-used baseline models: DeepLabV3+ \cite{chen2017rethinking} and UperNet \cite{xiao2018unified}. We'll highlight the advancements of our model in addressing challenges previously outlined.

\noindent
\underline{\textit{Small objects}}: 
Fig. \ref{fig:solar_pv_comparison} a) and c) highlight \model's proficiency in identifying small objects, even as diminutive as solar cells measuring $10 \times 5$ pixels, underscoring its potential in high-resolution aerial imagery.
\noindent
\underline{\textit{Intra-class heterogeneity}}:
Figs.\ref{fig:solar_pv_comparison} a), b), c) illustrate the diverse PV installations even within a single rooftop. The variety in shapes, colors, and orientations poses challenges for deep learning. Notably, \model accurately maps varied color cells in Fig. \ref{fig:solar_pv_comparison} a) and identifies accurate boundary in Fig. \ref{fig:solar_pv_comparison} b).


\noindent
\underline{\textit{Inter-class heterogeneity}}: Figs.\ref{fig:solar_pv_comparison} b), c) depict challenges due to visual similarities between PV installations and other elements. In Fig. \ref{fig:solar_pv_comparison} b), while both DeepLabV3+ and UperNet falter due to the rooftop's similar tone to PV installations, \model excels. In Fig. \ref{fig:solar_pv_comparison} c), DeepLabV3+ erroneously misidentifies a pool reflection as a solar cell. 

\noindent
\underline{\textit{Overall performance}}: Fig. \ref{fig:solar_pv_comparison} exhibits \model's superior ability to accurately depict PV installations, outperforming baseline models in clarity and precision. It identifies minute panels throughout the image and consistently offers robust results with minimal background confusion.

\section{Conclusion}

In this research, we introduced \model, an innovative Transformer-based model for solar PV profiling. Leveraging the Transformer architecture, \model adeptly learns high-resolution features and attend to the finer details while retaining complex relationships between them. This not only empowers \model with a comprehensive context but also lets it extract rich visual cues, as underscored by the robust results achieved. Thus, \model is able to address the solar PV challenges including low-resolution, inter-/intra-class homogenity. 

We benchmarked \model against current SOTA Solar-PV profiling methods and evaluated its performance alongside various CNN-based and Transformer-based segmentation networks using both ResNet-50 and ResNet-101 backbones. Our comprehensive quantitative analysis confirms that \model surpasses the results of other prevalent segmentation networks. Furthermore, our qualitative findings underscore its enhancements over prior methodologies.

\footnotesize
\bibliographystyle{IEEEtran}
\bibliography{conference/conference_main}

\vfill

\end{document}